\documentclass[conference]{IEEEtran}
\IEEEoverridecommandlockouts
\usepackage{cmap}
\usepackage[T1]{fontenc}
\usepackage[utf8]{inputenc}
\usepackage{newtxtext,newtxmath}
\input{glyphtounicode}
\usepackage{microtype} % 安全的字偶距微调
\usepackage[hidelinks,bookmarks=false]{hyperref}  % 尽量靠后加载（这里已足够靠后）
\usepackage{etoolbox}
\AtBeginEnvironment{thebibliography}{\linespread{1.15}\let\footnotesize\small\small\selectfont}

\usepackage{cite}                 % 不用 noadjust

\usepackage{amsmath,amssymb,amsfonts}
\usepackage{graphicx}
\usepackage{import}               % 用于 Inkscape PDF+LaTeX (.pdf_tex)
\usepackage{xcolor}
\usepackage[caption=false,font=footnotesize]{subfig}
\usepackage{booktabs,multirow,tabularx,array}
\usepackage{makecell}
\usepackage{dblfloatfix}         % fix ordering of double-column floats vs single-column
\usepackage[ruled,vlined,linesnumbered]{algorithm2e}

\DontPrintSemicolon
\SetKwInput{KwRequire}{Require}
\SetKwInput{KwEnsure}{Ensure}   % 只保留一次，避免重复定义
\SetKw{KwDownTo}{down to}
\SetAlFnt{\small}
\SetAlCapFnt{\small}
\SetAlCapNameFnt{\small}
\SetAlgoSkip{0.5em}

\def\BibTeX{{\rm B\kern-.05em{\sc i\kern-.025em b}\kern-.08em
    T\kern-.1667em\lower.7ex\hbox{E}\kern-.125emX}}
    
\begin{document}

\title{ComVLA: Communication-Aware Split Inference for VLA Models in 6G-Connected Robotics \\
}

\author{
    \IEEEauthorblockN{Boliang Liu\IEEEauthorrefmark{1}\IEEEauthorrefmark{2},
        Wint Yi Poe\IEEEauthorrefmark{2},
        Jingyun Di\IEEEauthorrefmark{1}\IEEEauthorrefmark{2},
        Riccardo Trivisonno\IEEEauthorrefmark{2}
        and Giuseppe Caire\IEEEauthorrefmark{1}}
    \IEEEauthorblockA{\IEEEauthorrefmark{1}Technical University of Berlin, Berlin, Germany\\
        % \IEEEauthorrefmark{2}Huawei Technologies D{\"u}sseldorf GmbH, Heisenberg Research Center, Munich, Germany\\
        \IEEEauthorrefmark{2}Huawei Heisenberg Research Center, Munich, Germany\\
        Emails: boliang.liu@campus.tu-berlin.de;\; \{wint.yi.poe, jingyun.di, riccardo.trivisonno\}@huawei.com;\; caire@tu-berlin.de}
}

\maketitle

\begin{abstract}
    Connected robotics is an emerging 6G application where mobile robots follow natural-language instructions to manipulate physical objects. The Vision-Language-Action (VLA) models that enable this are too large to run on the robot; a common trend is to offload inference to the cloud. The wireless link, however, limits how much sensing data the edge can transmit per control step. Two recent lines address this constraint: semantic communication codecs compress sensor data but require channel-specific retraining, and VLA token pruners select tokens from image but ignore the channel. Our insight is that the dense semantic information contained in the language already indicates which visual tokens matter. We propose ComVLA, a framework that uses this language guidance to adapt the VLA token budget to the channel capacity.
    Transmitting 32 tokens instead of 512 on the LIBERO benchmark, ComVLA cuts inference compute by $74\%$ and inference latency by $22\%$ versus the original OpenVLA-OFT baseline, at a cost of $1.5$\,pp in average task success ($95.4\%$ vs.\ $96.9\%$), and it stays within the capacity budget under Rayleigh and Rician fading. These results demonstrate that co-designing VLA inference and wireless communication is a practical direction for 6G-connected robotics.
\end{abstract}


\section{Introduction}
%1.5 pages
\label{sec:intro}

% Background & Motivation
% 机器人作为未来6G网络的关键应用，随着大语言模型（LLM）与视觉-语言模型（VLM）的飞速发展催生了机器人领域的范式转换。器人控制系统目前正经历由从传统的“感知-规划-控制”孤立模块化架构向视觉-语言-动作（VLA）模型驱动的重大转型：这是一种将摄像头、语言和动作信息进行端到端处理的多模态基础模型。
Connected robotics is among the demanding applications envisioned for 6G networks \cite{9040264}, requiring low-latency, high-reliability communication between mobile robots and cloud infrastructure. Recent advances in large language models (LLMs) and vision-language models (VLMs) have driven a paradigm shift in robotics, from traditional perception-planning-control pipelines toward Vision-Language-Action (VLA) models. VLAs extend VLMs with an action head, mapping camera observations and natural-language instructions to low-level actions end-to-end.

\begin{figure}[ht!]
    \centering
    \includegraphics[width=0.72\columnwidth]{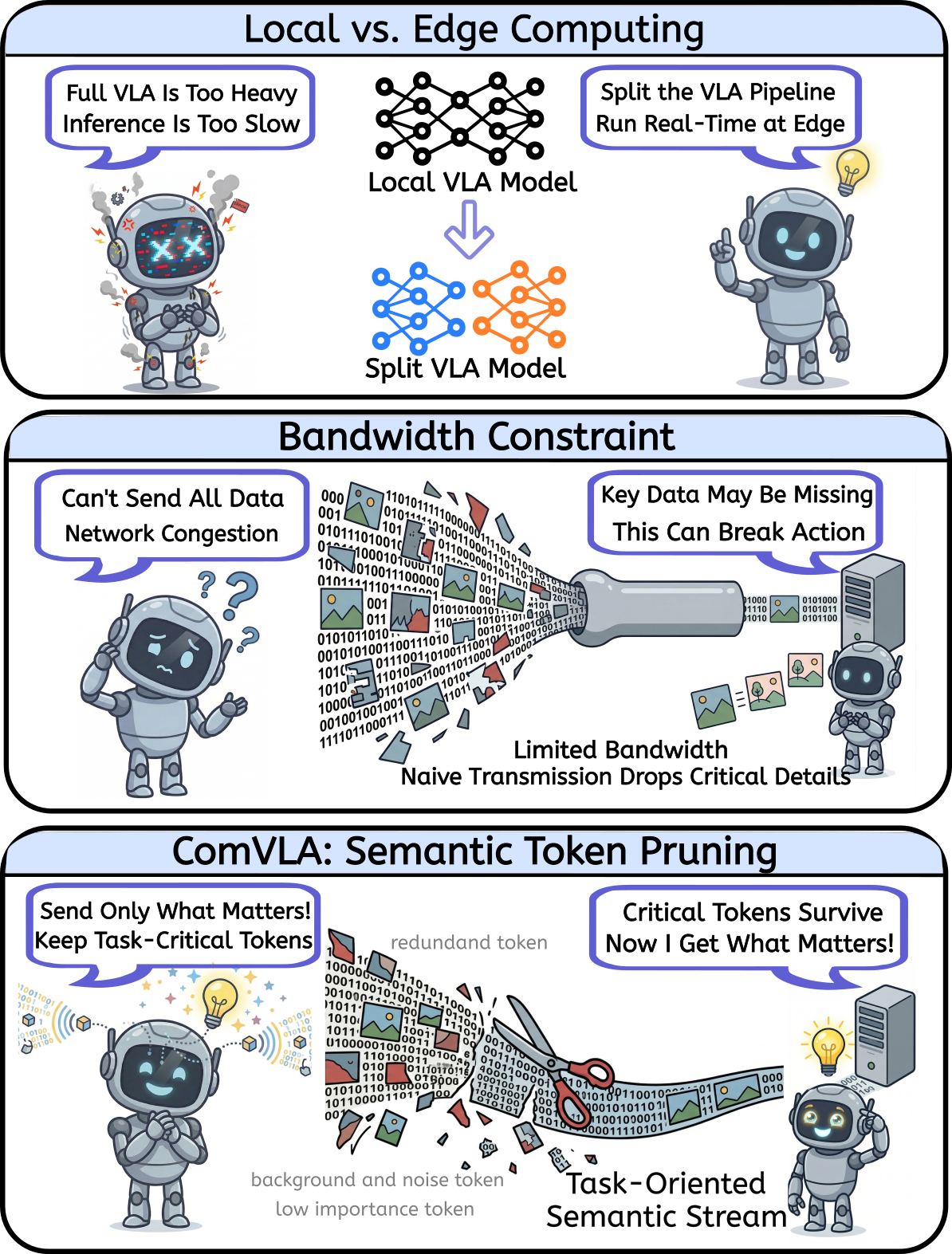}
    \caption{The motivation of ComVLA. Running full VLA models locally on edge robots is computationally prohibitive, while streaming all visual features to the cloud causes severe bandwidth bottlenecks. ComVLA transmits only the tokens that cross-attention voting identifies as task-critical, reducing both bandwidth and cloud compute.}
    \label{fig:motivation}
\end{figure}

The large VLA parameter scale makes edge deployment impractical, motivating split inference across edge and cloud \cite{chen2026goalorientedcommunicationfastrobust,ye10436772,ye2026selectthinkunlockingslm}. However, cloud offloading requires streaming high-dimensional visual features at each control step, which causes severe bandwidth bottlenecks over practical wireless channels. Yet recent findings identify a remedy: visual tokens are highly redundant in large vision-language models, and VLA action decisions depend on a sparse set of task-relevant visual regions~\cite{jiang2025betterlearnsmarterprune,Bi2025PRISMSI}.

Two lines of work exploit this sparsity but each leaves a gap. Semantic communication (SemCom)~\cite{xie2021deep} addresses the bandwidth mismatch by transmitting task-relevant features rather than raw data, but existing SemCom codecs optimize pixel reconstruction rather than task success and require channel-specific retraining. Beyond SemCom, recent VLA token pruning methods~\cite{pei2025vlaadp} reduce inference computing by selecting visual tokens via language instructions, but assume a static token budget with no mechanism to adapt to dynamic wireless conditions.

In this paper, we propose ComVLA (Communication-aware VLA), a split-inference framework that co-designs semantic token selection with the wireless channel, as illustrated in \figurename~\ref{fig:motivation}. Our key observation is that language instructions naturally identify the task-relevant visual tokens. ComVLA exploits this on the edge and dynamically sets the transmitted token budget to match the instantaneous channel capacity ~\cite{di11587624}.

%Experimental Results & Contributions
% 具体而言，本文的主要贡献总结如下：

% 新型端云协同范式 (ComVLA)： 我们提出了 ComVLA 框架，首次将 VLA 模型与通信设计深度融合。通过利用 VLA 固有的稀疏性，该框架将原本不感知通信的机器人控制转化为面向 6G 网络的通信感知新范式。

% 任务驱动的自适应剪枝： 我们设计了一种轻量级的容量约束语义剪枝机制。它能根据任务重要性动态过滤视觉 Token，使传输负载严格适配通信预算，从而大幅降低带宽开销与边缘计算延迟。

% 卓越的实验性能与鲁棒性： 在 LIBERO 基准测试上的广泛评估表明，ComVLA 实现了 96.5% 的平均成功率，并在极限带宽约束下依然保持了 95.4% 的强健表现，充分验证了其在边缘部署中的巨大潜力。
The contributions of this paper are:
\begin{itemize}
    \item \textbf{Framework - A communication-aware VLA inference paradigm.} We propose ComVLA, the first split-inference framework for VLA-based robotic manipulation that frames token pruning as an information-theoretically motivated, capacity-constrained token-selection problem, establishing a concrete co-design surface between VLA inference and the wireless link for 6G-connected robotics.
    \item \textbf{Mechanism - Semantic-guided token selection with a capacity-aligned token budget.} We use language instructions to evaluate the task relevance of visual tokens and couple the transmitted token budget directly to the channel capacity, so the payload adapts to the channel in real time.
    \item \textbf{Results - Robust performance under static and dynamic channels.} Transmitting 32 instead of 512 tokens, ComVLA cuts cloud compute by $3.8\times$ and inference latency by $22\%$ at a cost of only $1.5$\,pp in task success rate, and maintains robust performance under dynamic fading channels.
\end{itemize}

\section{Related Work}
\label{sec:related}

\textbf{Semantic and Task-Oriented Communication.} DeepJSCC~\cite{bourtsoulatze2019deep} pioneered wireless image transmission, and WITT~\cite{yang2023witt} later extended it with transformers; a parallel line of task-oriented communication~\cite{xie2021deep} targets task-level semantic objectives rather than pixel fidelity. These approaches all introduce dedicated codecs that require channel-specific retraining.

\textbf{VLA Models in Edge-Cloud Systems.}
VLA models demonstrate strong robotic manipulation capabilities but are too large for on-device inference. Such scale motivates split-computing paradigms~\cite{liu11588119} that partition inference between edge and cloud, yet existing frameworks still transmit raw visual features over bandwidth-limited channels and treat the link as a fixed pipe. ComVLA co-designs the split-inference token payload with channel capacity.

\textbf{Visual Token Pruning.}
Visual token pruning methods reduce inference costs but are channel-agnostic. FastV~\cite{chen2024image} shows up to half of visual tokens become redundant after the early transformer layers. For VLAs, LightVLA~\cite{jiang2025betterlearnsmarterprune} uses cross-attention on the OpenVLA-OFT backbone, and VLA-ADP~\cite{pei2025vlaadp} further gates tokens by action dynamics; both, however, target compute reduction and leave the token count blind to channel state~\cite{bi-etal-2025-llava}. Cross-attention has become a standard language-conditioned scoring signal across VLA token-pruning methods; ComVLA differs in that the token budget is set by the channel capacity rather than by a fixed compute target.

\section{System Model and Communication-Constrained Token Selection}
\label{sec:design}

\begin{figure*}[!t]
    \centering
    \includegraphics[width=0.88\textwidth]{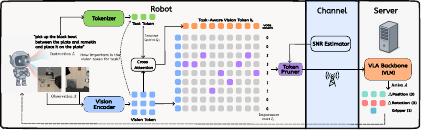}
    \caption{Overview of the ComVLA system. The edge robot extracts visual tokens from raw observations and estimates per-token task importance via language-visual cross-attention: language queries attend over visual tokens to form task-aware representations, which are then matched back to the original visual tokens by similarity to assign importance scores via hard voting. Only the top-$K^*$ tokens, selected according to the channel capacity, are transmitted to the cloud server for VLA inference and action prediction.}
    \label{fig:system}
\end{figure*}

The system operates in a split-inference configuration (\figurename~\ref{fig:system}). At each control step, the VLA's text tokenizer maps the instruction $L$ to language queries $Q_L$ and the visual encoder maps the observation $X$ to the full token set $\mathbf{V}=\{v_1,\ldots,v_N\}$. The ComVLA TokenPruner scores each visual token via a lightweight cross-attention pass (detailed in §\ref{sec:scoring}) and retains the top-$K^*$ tokens, where $K^*$ is set from the channel-state-information (CSI)-estimated link capacity of the current control period. The transmitted payload $Z$ is this pruned subset of visual tokens. Upon reception, the cloud server feeds the payload $\hat{Z}$ into the VLA backbone to generate the robotic action $A$. Each token retains its position encoding, and visual tokens are spatially redundant, so a sparse subset preserves action-relevant evidence.

The wireless channel imposes a capacity constraint
\begin{equation}
    \label{eq:link_capacity}
    C_{link} = \eta \cdot W \cdot T \cdot \log_2(1 + \text{SNR}),
\end{equation}
where $W$, $T$, and $\text{SNR}$ denote the bandwidth, control period, and signal-to-noise ratio, respectively. The factor $\eta \in (0,1)$ denotes a system efficiency factor that converts the ideal capacity into effective token-transmission goodput. We assume SNR is estimated from the channel and fed back to the edge once per control period. The language instruction $L$ and visual token position indices add negligible overhead.

We model the system as a Markov chain $X \xrightarrow{f_{\text{edge}}} Z \xrightarrow{\text{Channel}} \hat{Z} \xrightarrow{f_{\text{cloud}}} A$. This naturally frames token pruning as an Information Bottleneck (IB) problem. While classical IB balances compression and relevance to a target $Y$ via an unconstrained Lagrangian $\mathcal{L}_{\text{IB}} = I(Z;X) - \beta I(Z;Y)$, our robotics setup uses the language-conditioned action $A$ as the relevance target and enforces a strict effective link-budget constraint $C_{link}$:

\begin{equation*}
    \max_{Z} I(Z;A \mid L) \quad \text{s.t.} \quad I(Z;X \mid L) \le C_{link}.
\end{equation*}

We operationalize this constrained mutual information (MI) view through a tractable capacity-constrained token-selection objective. (Direct estimation of these MI terms is provably hard in high dimensions~\cite{pmlr-v108-mcallester20a}, and impractical at robot control rates.) By the entropy decomposition of conditional mutual information,
\begin{equation}
    \label{eq:mi_decomp}
    I(Z;A\mid L) = H(A\mid L) - H(A\mid Z,L),
\end{equation}
and since $H(A\mid L)$ does not depend on $Z$, the operational objective is the residual action uncertainty $H(A\mid Z,L)$ under the channel budget.

We assign each visual token a non-negative importance score $\mathcal{I}_j$, interpreted as an operational proxy for language-conditioned action relevance, and select a binary mask $m$ maximizing accumulated importance under the communication budget:

\begin{equation*}
    \max_{m \in \{0,1\}^N} \sum_{j=1}^N \mathcal{I}_j m_j \quad \text{s.t.} \quad \sum_{j=1}^N D \cdot B \cdot m_j \le C_{link}
\end{equation*}

where $m_j \in \{0,1\}$ indicates whether the $j$-th token is transmitted, and $D \cdot B$ is the bit overhead per visual token. Since all visual tokens have the same payload $D \cdot B$, the exact solution is to select the top-$K^*$ tokens with the largest importance scores, where $K^* = \lfloor C_{link}/(D \cdot B) \rfloor$. This decouples the edge problem into a channel-driven budget $K^*$ and a task-driven ranker $\{\mathcal{I}_j\}$. Section~\ref{sec:scoring} specifies our $\mathcal{I}_j$; Section~\ref{sec:exp} validates it on LIBERO.

\section{Channel-Adaptive Token Selection}
\label{sec:framework}

\subsection{Attention-Based Semantic Importance Estimation}
\label{sec:scoring}
We compute the importance score $\mathcal{I}_j$ via a cross-attention pass. The scoring runs entirely on the edge at an $O(|Q_L|{\times}N{\times}D)$ cost with no new trainable parameters. The VLA's text embedding layer produces language queries $Q_L \in \mathbb{R}^{|Q_L| \times D}$, which cross-attend over all visual tokens $\mathbf{V} = \{v_1, \ldots, v_N\}$ to form task-aware visual representations $h_i = \sum_j \alpha_{i,j} v_j$, where $\alpha_{i,j} = \operatorname{softmax}(q_i^\top \mathbf{V}/\sqrt{D})_j$. Each task-aware token $h_i$ is then matched to its most similar original visual token via dot-product similarity, and that token receives one hard vote; this argmax voting is compared against soft voting in Section~\ref{sec:voting}. The task-importance score $\mathcal{I}_j$ counts the votes received:

\begin{equation}
    \label{eq:importance}
    \mathcal{I}_j = \left|\left\{ i \;\middle|\; j = \mathop{\arg\max}_{k}\, h_i^\top v_k \right\}\right|
\end{equation}

This argmax-based selection aligns with the IB view of §\ref{sec:design}: tokens that are no query's argmax contribute no estimable reduction in $H(A\mid Z,L)$ and can be discarded.

\subsection{Capacity-Constrained Token Pruning}
The number of selected tokens $K^*$ must match the link budget. The capacity-aligned token budget is:

\begin{equation}
    \label{eq:kstar}
    K^*(\text{SNR}) = \left\lfloor \frac{\eta \cdot W \cdot T \cdot \log_2(1 + \text{SNR})}{D \cdot B} \right\rfloor
\end{equation}

$K^*$ is the per-step token budget supported by the effective goodput; Algorithm~\ref{alg:comvla} summarizes the procedure.

\begin{figure*}[!t]
    \centering
    \includegraphics[width=0.72\textwidth]{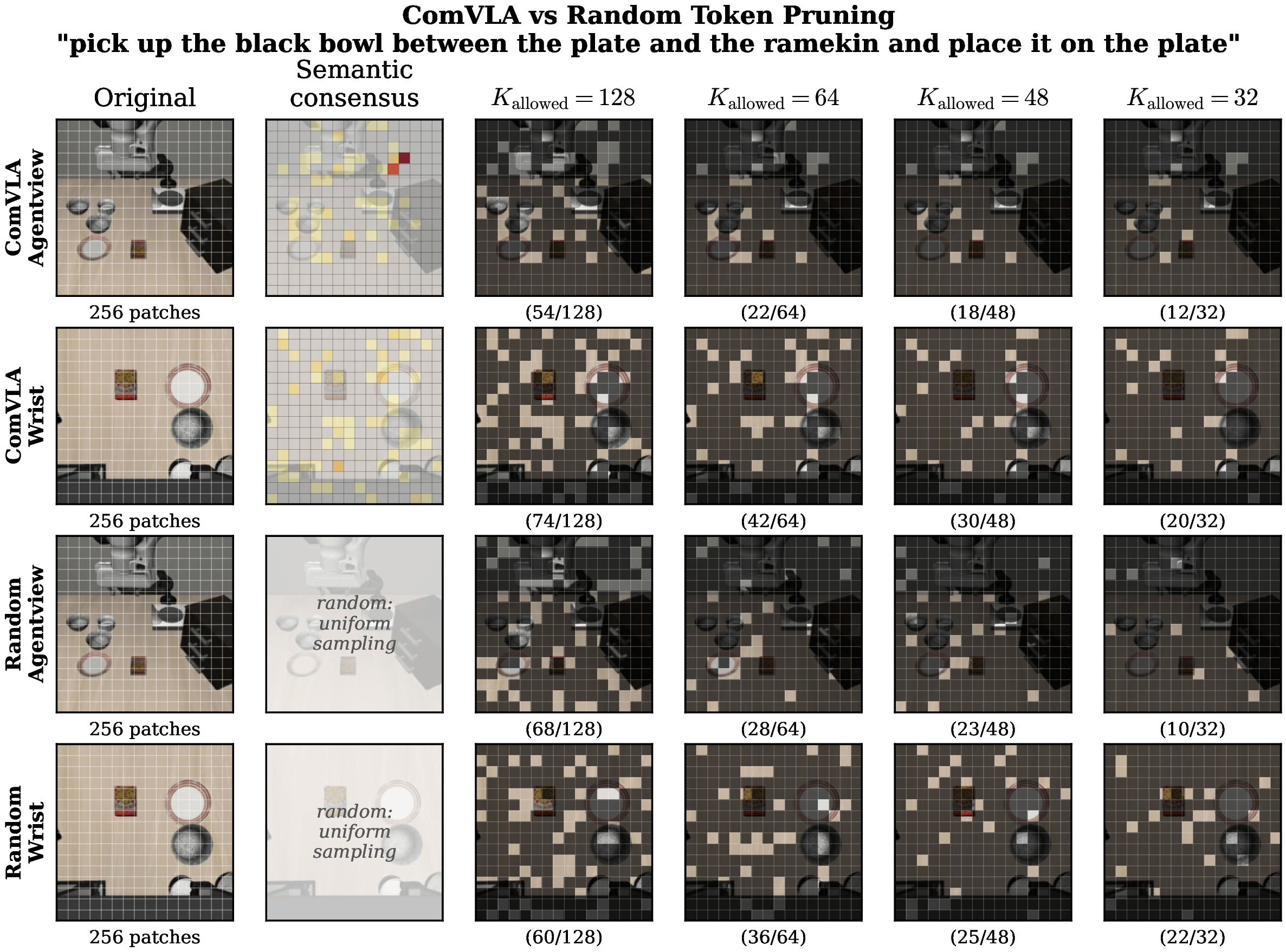}
    \caption{ComVLA vs.\ Random Token Pruning for a representative manipulation task from the LIBERO benchmark. ComVLA retains task-critical regions (e.g., the black bowl, the plate, and the end-effector) under varying bandwidth constraints ($K^* \in \{128, 64, 48, 32\}$). The random baseline uniformly samples patches, breaking the spatial continuity needed for control. Across all four LIBERO suites, ComVLA consistently concentrates selection on the end-effector, target object, and goal region; this example illustrates the general behavior. Each control step uses two camera views; $K^*$ is the total transmitted token count across both views.}
    \label{fig:token_pruning}
\end{figure*}

\begin{algorithm}[tb]
    \caption{Capacity-Constrained Semantic Pruning (ComVLA)}
    \label{alg:comvla}
    \LinesNotNumbered
    \KwRequire{Raw observation $X$, Language instruction $L$, Channel bandwidth $W$, Period $T$, Feature dimension $D$, Quantization $B$, Current $\text{SNR}$.}
    \KwEnsure{Pruned token subset $Z$.}
    \tcc{Edge: Visual Encoding}
    Extract full token set $\mathbf{V} = \{v_1, \dots, v_N\}$ from $X$ via the visual encoder\;
    \tcc{Edge: Language Embedding (text tokenizer + embedding layer only)}
    $Q_L \leftarrow \text{TextEmbed}(L)$ \tcp*{no VLM decoding; produces $|Q_L|$ query vectors}
    Initialize importance scores $\mathcal{I}_j = 0$ for all $v_j \in \mathbf{V}$\;
    \tcc{Edge: Cross-Attention Scoring Pass (ranking only; no action head)}
    \ForEach{query $q_i \in Q_L$}{
        Compute attention weights: $\alpha_{i,j} = \operatorname{softmax}(q_i^\top \mathbf{V}/\sqrt{D})_j$\;
        Form task-aware token: $h_i = \sum_j \alpha_{i,j} v_j$\;
        Match to closest visual token: $j^* = \mathop{\arg\max}_{k}\, h_i^\top v_k$\;
        Hard voting: $\mathcal{I}_{j^*} \leftarrow \mathcal{I}_{j^*} + 1$\;
    }
    \tcc{Physical Channel Alignment}
    Calculate effective-goodput token budget $K^*(\text{SNR})$ via Eq.~\eqref{eq:kstar}\;
    \tcc{Capacity-Constrained Token Pruning}
    Sort $\mathbf{V}$ in descending order based on scores $\mathcal{I}$\;
    Select the top-$K^*$ tokens to form the transmission payload $Z$\;
    \Return $Z$\;
\end{algorithm}

As shown in \figurename~\ref{fig:token_pruning} for a representative task from the LIBERO benchmark, ComVLA concentrates the limited budget on task-critical regions: the end-effector, the target object, and the goal container, across $K^* \in \{128, 64, 48, 32\}$, whereas random pruning scatters kept tokens uniformly and loses the spatial continuity around the manipulated object.

\section{Performance Evaluation}
\label{sec:exp}

We evaluate ComVLA on the LIBERO benchmark~\cite{liu2023libero}, which comprises four task suites (Spatial, Object, Goal, and Long Horizon), each with 10 tasks evaluated over 50 episodes. The base model is OpenVLA-OFT~\cite{kim2025openvlaoft} fine-tuned on LIBERO~\cite{jiang2025betterlearnsmarterprune}; we use the Section~\ref{sec:design} pipeline with SigLIP and DINOv2 visual encoders on the edge. We compare against: \textbf{OpenVLA-OFT} with the full 512-token input; \textbf{LightVLA}~\cite{jiang2025betterlearnsmarterprune}, a content-adaptive cross-attention pruner; \textbf{VLA-ADP}~\cite{pei2025vlaadp}, which additionally gates tokens by action dynamics; \textbf{DeepJSCC}~\cite{bourtsoulatze2019deep} and \textbf{WITT}~\cite{yang2023witt}, which operate on reconstructed pixel space and therefore still require full OpenVLA-OFT at the cloud; and \textbf{Random Pruning}, which selects $K^*$ tokens uniformly at random. All experiments run on H100 GPU.

\textbf{TX Data accounting.} Token-based methods (OpenVLA-OFT, LightVLA, VLA-ADP, ComVLA, Random Pruning) transmit $\mathrm{TX} = K \cdot D \cdot B / 8$ bytes per control step, where $K$ is the transmitted token count, $D$ is the token feature dimension, and $B$ is the per-feature bit-depth from Eq.~\eqref{eq:kstar} under the chosen quantization scheme (analyzed in Section~\ref{sec:quantization}). DeepJSCC and WITT transmit the codec's real-valued output latent $\mathbf{z}\in\mathbb{R}^{C_\ell\times H_\ell\times W_\ell}$ in full, giving $\mathrm{TX} = |\mathbf{z}| \cdot B_\ell / 8$ bytes at the codec's native per-scalar bit-width $B_\ell$. The fixed-$K$ settings in Table~\ref{tab:main_comparison} represent capacity-equivalent token budgets induced by Eq.~\eqref{eq:kstar}, ranging from moderate communication constraints to extreme low-goodput stress testing; the SNR-driven $K^*$ mapping is evaluated in §\ref{sec:fading}.

\textbf{FLOPs and latency accounting.} Table~\ref{tab:flops_latency} breaks down FLOPs and compute latency by component. All reported latencies refer to compute-side inference only. Under 6G's sub-millisecond air-interface targets~\cite{9040264,}, wireless transmission contributes negligibly relative to the compute latency reported here. Following standard analytical FLOPs counting~\cite{korthikanti2023reducing}, for each transformer layer:
\begin{equation}
    \label{eq:flops}
    C_{\text{layer}} = \underbrace{8sd^2}_{\text{QKV + out proj.}} + \underbrace{4s^2d}_{\text{attention}} + \underbrace{6sdf}_{\text{FFN (SwiGLU)}}
\end{equation}
The quadratic $4s^2d$ attention term means that reducing the token count has a disproportionate effect on VLM cost.

\subsection{Task Success and Compute Efficiency}
\begin{table*}[!t]
    \centering
    \caption{Comparison of split-inference methods on LIBERO benchmarks (50 episodes $\times$ 10 tasks per suite, INT8 token accounting). ComVLA achieves comparable task success to unconstrained baselines ($\leq1.5$\,pp gap) while dynamically adapting $K^*$ to channel capacity.}
    \label{tab:main_comparison}
    \setlength{\tabcolsep}{4.5pt}
    \renewcommand{\arraystretch}{1.15}
    \begin{tabular}{ll cccc cccc c}
        \toprule
        \multirow{2}{*}{\textbf{Method}}                 & \multirow{2}{*}{\textbf{Config}} & \multirow{2}{*}{\makecell[c]{\textbf{TX Data $\downarrow$}                                                                                                                         \\\textbf{(KB/step)}}} & \multirow{2}{*}{\makecell[c]{\textbf{FLOPs $\downarrow$}\\\textbf{(TFLOPs)}}} & \multirow{2}{*}{\makecell[c]{\textbf{Latency $\downarrow$}\\\textbf{(ms)}}} & \multicolumn{4}{c}{\textbf{Success Rate (\%) $\uparrow$}} & \multirow{2}{*}{\textbf{Avg.\ (\%) $\uparrow$}} \\
        \cmidrule(lr){6-9}
                                                         &                                  &                                                            &                  &               & \textbf{Spatial} & \textbf{Object} & \textbf{Goal} & \textbf{Long} &               \\
        \midrule
        OpenVLA-OFT~\cite{kim2025openvlaoft}             & 512 tokens                       & 1088.0                                                     & 8.8              & 54.8          & 98.4             & 97.0            & 97.0          & 95.2          & 96.9          \\
        VLA-ADP~\cite{pei2025vlaadp}                     & dynamic                          & 920.4                                                      & 7.7              & 52.5          & 92.4             & 98.4            & 96.8          & 94.4          & 95.5          \\
        LightVLA~\cite{jiang2025betterlearnsmarterprune} & dynamic                          & 165.3                                                      & 2.9              & 56.9          & 98.6             & 97.8            & 98.0          & 94.6          & 97.3          \\
        \midrule
        DeepJSCC~\cite{bourtsoulatze2019deep}            & unconstrained                    & 392.0                                                      & 8.8$^{\ddagger}$ & 55.5          & 98.2             & 96.6            & 97.8          & 92.4          & 96.3          \\
        WITT~\cite{yang2023witt}                         & unconstrained                    & 192.0                                                      & 8.9$^{\ddagger}$ & 67.6$^{\S}$   & 97.8             & 97.8            & 97.8          & 93.8          & 96.8          \\
        \midrule
        Random Prune                                     & $K{=}128$                        & 272.0                                                      & 2.9              & 42.5          & 96.8             & 96.0            & 96.0          & 91.6          & 95.1          \\
        Random Prune                                     & $K{=}64$                         & 136.0                                                      & 2.7              & 42.6          & 94.4             & 93.0            & 96.0          & 84.2          & 91.9          \\
        Random Prune                                     & $K{=}48$                         & 102.0                                                      & 2.5              & 42.6          & 87.4             & 83.8            & 94.2          & 74.0          & 84.9          \\
        Random Prune                                     & $K{=}32$                         & 68.0                                                       & 2.3              & 42.7          & 65.8             & 32.6            & 79.6          & 44.8          & 55.7          \\
        Random Prune                                     & $K{=}16$                         & 34.0                                                       & 2.1              & 42.9          & 8.0              & 0.0             & 22.8          & 0.0           & 7.7           \\
        \midrule
        \textbf{ComVLA}                                  & $\boldsymbol{K{=}128}$           & \textbf{272.0}                                             & \textbf{2.9}     & \textbf{42.4} & \textbf{98.2}    & \textbf{96.6}   & \textbf{96.4} & \textbf{92.2} & \textbf{95.9} \\
        \textbf{ComVLA}                                  & $\boldsymbol{K{=}64}$            & \textbf{136.0}                                             & \textbf{2.7}     & \textbf{42.6} & \textbf{97.8}    & \textbf{98.0}   & \textbf{96.8} & \textbf{93.0} & \textbf{96.4} \\
        \textbf{ComVLA}                                  & $\boldsymbol{K{=}48}$            & \textbf{102.0}                                             & \textbf{2.5}     & \textbf{42.3} & \textbf{97.2}    & \textbf{98.2}   & \textbf{97.0} & \textbf{91.2} & \textbf{96.0} \\
        \textbf{ComVLA}                                  & $\boldsymbol{K{=}32}$            & \textbf{68.0}                                              & \textbf{2.3}     & \textbf{42.7} & \textbf{96.2}    & \textbf{97.0}   & \textbf{97.6} & \textbf{90.8} & \textbf{95.4} \\
        \textbf{ComVLA}                                  & $\boldsymbol{K{=}16}$            & \textbf{34.0}                                              & \textbf{2.1}     & \textbf{42.7} & \textbf{37.4}    & \textbf{0.2}    & \textbf{64.4} & \textbf{35.6} & \textbf{34.4} \\
        \bottomrule
    \end{tabular}
    \vspace{2pt}
    \par\raggedright\footnotesize
    $^{\ddagger}$ WITT/DeepJSCC FLOPs include codec encoder/decoder (WITT: 0.1\,T, DeepJSCC: $<$0.01\,T) + full OpenVLA-OFT inference (8.8\,T). \quad
    $^{\S}$ WITT latency includes 12.8\,ms codec overhead.
\end{table*}

As shown in Table~\ref{tab:main_comparison}, existing pruners like LightVLA and VLA-ADP achieve high success rates (97.3\% and 95.5\%) using content-adaptive token budgets. Their token-importance heads are trained at a fixed budget with no runtime mechanism for adjusting $K$; ComVLA addresses this by deriving $K^*$ from the channel SNR at each control step (Eq.~\eqref{eq:kstar}). Forcing them below the trained budget drops tokens the importance head can't rank, reducing to random dropping; Table~\ref{tab:main_comparison}'s Random Pruning rows stand in for that regime. The pixel-level codecs stay competitive on success rate but do not reduce cloud compute or token count: they still run full OpenVLA-OFT inference (8.8\,TFLOPs) and add codec latency. Random Pruning degrades sharply under tight budgets, falling to 55.7\% at $K^*{=}32$ and collapsing to 32.6\% on the Object suite, a 64.4\,pp gap versus ComVLA's 97.0\%. ComVLA reaches 95.4\% at $K^*{=}32$ (68\,KB/step), reducing transmission by $16{\times}$ and cloud compute from 8.8 to 2.3\,TFLOPs.

Long Horizon's larger drop ($-4.4$\,pp at $K^*=32$) tracks its higher baseline difficulty (95.2\% vs.\ 97--98\% on the others). $K{=}16$ pushes both ComVLA and Random Pruning beyond the practical operating range.

\begin{table}[ht!]
    \centering
    \caption{Per-component FLOPs and latency for OpenVLA-OFT vs.\ ComVLA ($K^*=32$).}
    \label{tab:flops_latency}
    \setlength{\tabcolsep}{2pt}
    \renewcommand{\arraystretch}{1.25}
    \begin{tabular}{l cc cc cc}
        \toprule
        \multirow{2}{*}{\textbf{Method}}
                        & \multicolumn{2}{c}{\textbf{Vision Enc.$^\dagger$}}
                        & \multicolumn{2}{c}{\textbf{VLM}}
                        & \multicolumn{2}{c}{\textbf{System Total}}                                                                                              \\
        \cmidrule(lr){2-3}\cmidrule(lr){4-5}\cmidrule(lr){6-7}
                        & \textbf{TFLOPs (TF)}                               & \textbf{Lat.(ms)}
                        & \textbf{TF}                                        & \textbf{Lat.(ms)}
                        & \textbf{TF}                                        & \textbf{Lat.(ms)}                                                                 \\
        \midrule
        OpenVLA-OFT     & 0.83                                               & 22.6              & 7.97          & 32.2          & 8.80          & 54.8          \\
        \textbf{ComVLA} & \textbf{0.83}                                      & \textbf{22.6}     & \textbf{1.46} & \textbf{20.1} & \textbf{2.29} & \textbf{42.7} \\
        \bottomrule
    \end{tabular}
\end{table}

As Table~\ref{tab:flops_latency} shows, the robot's on-device compute drops from 8.80\,TFLOPs (running the full OpenVLA-OFT on device) to 0.83\,TFLOPs (vision encoder only, $\downarrow 90.6\%$), as the VLM is fully offloaded to the cloud server. Cloud-side latency is dominated by autoregressive action generation, so compute savings from smaller $K^*$ do not translate into latency savings (42.3--42.7\,ms across all $K$). The practical payoff is TX data and cloud compute, which drop near-linearly with $K^*$ while inference latency holds steady.

\subsection{Performance under Fading-Aware Token-Budget Emulation}
\label{sec:fading}

% \textcolor{blue}{\subsection{Time-Varying Fading Channel Model}
% The wireless link is modeled as a simplified time-varying fading channel, where the instantaneous SNR is used as the channel-state indicator for transmission-budget adaptation:
% \begin{equation}
% \gamma(t)=\bar{\gamma}|h(t)|^2 .
% \end{equation}
% Two fading scenarios are considered: Rayleigh and Rician fading. Rayleigh fading represents a rich-scattering environment without a dominant line-of-sight path, where multipath superposition can cause large fluctuations in $|h(t)|^2$ and thus in the output SNR, even under a fixed average SNR. In contrast, Rician fading includes a stable line-of-sight component,
% \begin{equation}
% h_{\mathrm{Ric}}(t)
% =
% \sqrt{\frac{K}{K+1}}h_{\mathrm{LOS}}
% +
% \sqrt{\frac{1}{K+1}}h_{\mathrm{sc}}(t),
% \end{equation}
% which mitigates deep fading and leads to a smoother SNR trajectory. The resulting SNR is then used to estimate the available capacity and determine the effective transmission budget.}

The wireless link is modeled as a simplified time-varying fading channel with instantaneous SNR driving the transmission-budget adaptation. Two scenarios are considered: Rayleigh fading exhibits larger SNR fluctuations and deeper troughs, while Rician fading retains a stable line-of-sight component that yields a smoother SNR trajectory. We exclude fixed-budget pruners since they cannot adapt to tight capacities, and SemCom codecs because they require SNR-specific retraining.

\begin{figure}[ht!]
    \centering
    \includegraphics[width=0.92\columnwidth]{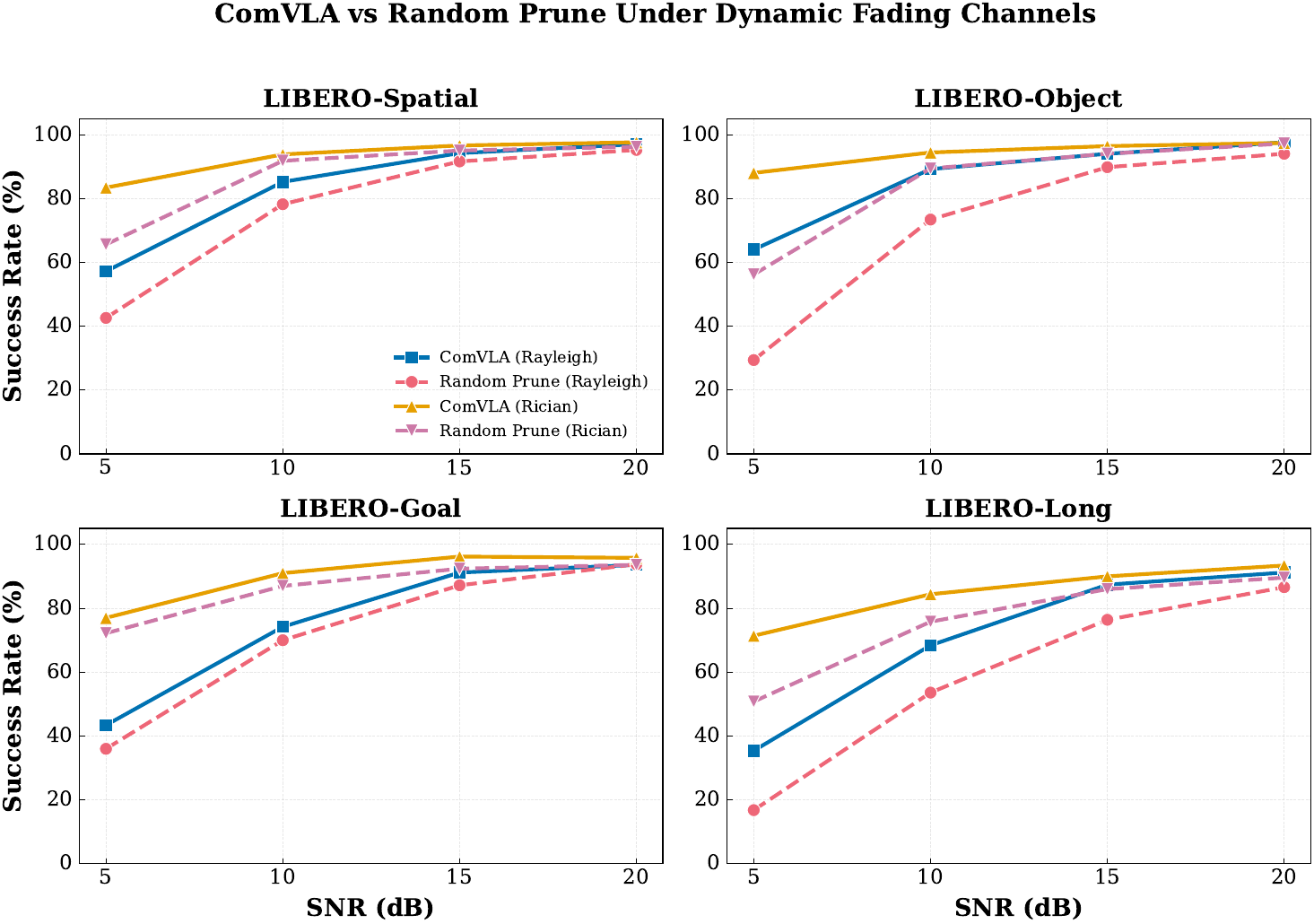}
    \caption{Success rate vs.\ SNR for ComVLA and Random Pruning under Rayleigh and Rician fading ($\kappa=3.0$) with $W= 20$\,MHz, $T=50$\,ms (robot VLA control period), $f_d=10$\,Hz. Both methods dynamically adjust the token budget $K^*$ based on our abstracted effective payload budget model. ComVLA's semantic-aware selection maintains higher success rates across all suites, with the largest gap at low SNR under Rayleigh fading.}
    \label{fig:dynamic_channel}
\end{figure}

As shown in \figurename~\ref{fig:dynamic_channel}, ComVLA degrades more gracefully than Random Pruning as SNR drops. On the Object suite at SNR\,=\,5\,dB under Rician fading, ComVLA reaches 88.0\% versus 56.2\% for Random Pruning (+31.8\,pp). Rician outperforms Rayleigh at the same average SNR: smaller SNR fluctuations avoid the deep troughs that transiently push $K^*$ into the steep-degradation seen at very low budgets in Table~\ref{tab:main_comparison}.

\begin{figure}[ht!]
    \centering
    \includegraphics[width=0.5\columnwidth]{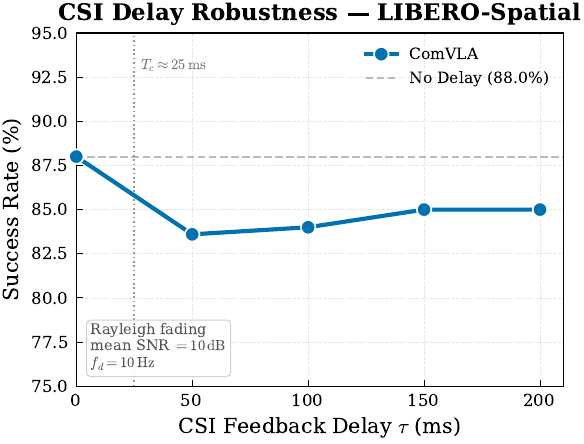}
    \caption{ComVLA success rate on LIBERO-Spatial under CSI feedback delay of $d\in\{0,\ldots,4\}$ control periods ($T=50$\,ms).}
    \label{fig:csi_delay}
\end{figure}

ComVLA is also robust to stale CSI: delaying the SNR estimate by $d\in\{0,\ldots,4\}$ control periods (up to 200\,ms) on LIBERO-Spatial yields only a marginal success-rate drop (\figurename~\ref{fig:csi_delay}), so instantaneous feedback is not required.

\subsection{Effect of Score Aggregation}
\label{sec:voting}
\begin{table}[ht!]
    \centering
    \caption{Hard vs.\ soft voting for ComVLA at $K^*=32$ (means over 500 episodes).}
    \label{tab:voting}
    \setlength{\tabcolsep}{4pt}
    \renewcommand{\arraystretch}{1.15}
    \begin{tabular}{l cccc c}
        \toprule
        \textbf{Voting}             & \textbf{Spatial} & \textbf{Object} & \textbf{Goal} & \textbf{Long} & \textbf{Avg.} \\
        \midrule
        Soft voting                 & 97.2             & 95.6            & 96.6          & 84.8          & 93.6          \\
        \textbf{Hard voting (ours)} & \textbf{96.6}    & \textbf{97.2}   & \textbf{97.6} & \textbf{89.8} & \textbf{95.3} \\
        \bottomrule
    \end{tabular}
\end{table}

We compare hard voting, defined in Eq.~\eqref{eq:importance}, against soft voting, which uses the attention weights directly as importance scores: $\mathcal{I}_j^{\text{soft}} = \sum_i \alpha_{i,j}$. Eq.~\eqref{eq:mi_decomp} implies that the ideal $\mathcal{I}_j$ should concentrate on the sparse set of tokens whose inclusion reduces $H(A\mid Z,L)$; under a tight budget $K^*$, any importance mass leaked to task-irrelevant patches directly displaces task-relevant ones from the top-$K^*$ selection. Because cross-attention weights are spread across many background tokens, soft voting leaks mass to background, whereas hard voting concentrates each query on its single foreground argmax. Table~\ref{tab:voting} shows an average gain of 1.7\,pp, driven by a +5.0\,pp jump on Long Horizon alongside a smaller regression on Spatial ($-0.6$\,pp); the effect is largest where the $K^*{=}32$ budget is tightest relative to task complexity. Connection to IB, hard voting's argmax commitment respects the IB-prescribed concentration of $\mathcal{I}_j$, which soft voting violates by spreading mass.

\subsection{Effect of Token Quantization}
\label{sec:quantization}

\begin{table}[ht!]
    \centering
    \caption{Effect of token quantization for ComVLA at $K^*=32$ (means over 500 episodes per suite). INT8 halves transmission cost over FP16 with negligible accuracy loss; INT4 causes severe degradation.}
    \label{tab:quantization}
    \setlength{\tabcolsep}{4pt}
    \renewcommand{\arraystretch}{1.15}
    \begin{tabular}{l c cccc c}
        \toprule
        \textbf{Quant.} & \textbf{TX Data} & \textbf{Spatial} & \textbf{Object} & \textbf{Goal} & \textbf{Long} & \textbf{Avg.} \\
        \midrule
        FP16            & 136              & 96.6             & 97.2            & 97.6          & 89.8          & 95.3          \\
        \textbf{INT8}   & \textbf{68}      & \textbf{96.2}    & \textbf{97.0}   & \textbf{97.6} & \textbf{90.8} & \textbf{95.4} \\
        INT4            & 34               & 69.8             & 73.8            & 75.8          & 49.8          & 67.3          \\
        \bottomrule
    \end{tabular}
\end{table}

Table~\ref{tab:quantization} shows that reducing precision from FP16 to INT8 at $K^*=32$ halves the transmission payload (136\,KB to 68\,KB) with negligible impact on task performance (95.3\%$\to$95.4\%). Further compression to INT4 causes substantial degradation (67.3\%): with only 16 discrete levels, fine-grained feature differences are clipped, corrupting the importance ordering. INT8 therefore balances bandwidth and accuracy.

\section{Conclusion}

We proposed ComVLA, a framework for bandwidth-adaptive split inference of VLA models. It uses language-guided vision token pruning as a transmission-priority signal, enabling the edge to select task-critical tokens. Transmitting only 32 of 512 tokens, ComVLA reduces per-step TX data by $16\times$ (1088\,KB to 68\,KB), cloud compute by $3.8\times$, and inference latency by $22\%$, at a cost of only $1.5$\,pp in task success. It remains effective under Rayleigh and Rician fading and tolerates stale CSI feedback up to $200$\,ms. The semantic structure already inside a VLA suffices for channel-adaptive transmission. Future work targets real-channel evaluation, real-world deployment, and uncertainty-guided token scoring.

\bibliographystyle{IEEEtran}
\bibliography{IEEEabrv,references}
\end{document}